\documentclass[conference]{IEEEtran}
\IEEEoverridecommandlockouts
\usepackage{cite}
\usepackage{amsmath,amssymb,amsfonts}
\usepackage{algorithmic}
\usepackage{graphicx}
\usepackage{textcomp}
\usepackage{xcolor}
\def\BibTeX{{\rm B\kern-.05em{\sc i\kern-.025em b}\kern-.08em
    T\kern-.1667em\lower.7ex\hbox{E}\kern-.125emX}}
\newcommand{\model}{DropMix}
\usepackage{amsfonts}
\usepackage{fancyhdr}
\usepackage{booktabs}
\usepackage{algorithm}
\usepackage{algorithmic}
\usepackage{amsfonts}
\usepackage{amsmath}
\usepackage{hyperref}
\makeatletter
\newcommand{\linebreakand}{%
  \end{@IEEEauthorhalign}
  \hfill\mbox{}\par
  \mbox{}\hfill\begin{@IEEEauthorhalign}
}
\def\thanks#1{\protected@xdef\@thanks{\@thanks
        \protect\footnotetext{#1}}}
\makeatother

\begin{document}

\title{DropMix: Better Graph Contrastive Learning
with Harder Negative Samples
}

\author{\IEEEauthorblockN{Yueqi Ma}
\IEEEauthorblockA{\textit{School of Data Science and Engineering} \\
\textit{East China Normal University}\\
Shanghai, China \\
51215903110@stu.ecnu.edu.cn}
\and
\IEEEauthorblockN{Minjie Chen}
\IEEEauthorblockA{\textit{School of Data Science and Engineering} \\
\textit{East China Normal University}\\
Shanghai, China \\
minjiechen@stu.ecnu.edu.cn}
\and
\IEEEauthorblockN{Xiang Li\IEEEauthorrefmark{1}}
\IEEEauthorblockA{\textit{School of Data Science and Engineering} \\
\textit{East China Normal University}\\
Shanghai, China \\
xiangli@dase.ecnu.edu.cn}
}

\thanks{\IEEEauthorrefmark{1} Corresponding author.}
\maketitle

\begin{abstract}
While generating better negative samples for contrastive learning has been widely studied in the areas of CV and NLP, very few work has focused on graph-structured data.
Recently,
Mixup has been introduced to synthesize
hard negative samples in graph contrastive learning (GCL).
However, 
due to the unsupervised learning nature of GCL,
without the help of soft labels,
directly mixing representations of samples could inadvertently lead to the information loss of the original hard negative and further adversely affect the quality of the newly generated harder negative.
To address the problem, 
in this paper,
we propose a novel method \model\ to synthesize harder negative samples, which consists of two main steps.
Specifically,
we first select some hard negative samples by measuring their hardness from both local and global views in the graph simultaneously. 
After that, 
we mix hard negatives only
on partial representation dimensions to generate harder ones and 
decrease the information loss caused by Mixup.  
We conduct extensive experiments to 
verify the effectiveness of \model\ on six benchmark datasets.
Our results show that our method can lead to better GCL performance. 
Our data and codes are publicly available at 
\url{https://github.com/Mayueq/DropMix-Code}.
\end{abstract}


\section{Introduction}
With the explosion of data volume in the real world, 
labeled data is generally hard to derive,
while 
more available data is unlabeled.
To leverage massive unlabeled data,
self-supervised learning (SSL) \cite{neville2000iterative,zhu2003semi}
has recently received increasing attention.
Contrastive learning (CL) \cite{bachman2019learning}, a mainstream type of SSL,
has been widely used in the fields of CV~\cite{he2020momentum,chen2020simple} and NLP~\cite{devlin2018bert,yang2019xlnet}.
The major goal of CL is to maximize the similarity between positive pairs while minimizing that between negative ones.
CL has also been extended to graphs 
and integrated with graph neural networks (GNNs)
to learn node/graph representations,
whose superior performance has been verified \cite{kipf2016semi}.

Some previous studies \cite{chen2020improved,tian2020makes} 
have shown that the key to affecting the performance of contrastive learning 
is the selection of positive and negative samples. 
Generally, 
positive samples are constructed by data augmentation methods~\cite{zhao2021data,you2020graph},
while negative samples are randomly selected.
Recently,
it has been pointed out in \cite{cai2020all} 
that harder negative samples are more helpful for learning.
Therefore,
there has been a growing interest in selecting harder 
negative samples for contrastive learning~\cite{robinson2020contrastive,wang2020reinforced}.
Further,
negative samples can also be
constructed by 
artificial synthesis.
For example,
Mixup~\cite{zhang2017mixup}
and 
CutMix~\cite{yun2019cutmix} are
two data augmentation methods that have been applied 
in the area of CV,
which mix different images and their labels to synthesize new samples with soft labels.
The difference is that 
Mixup performs 
interpolation between two images,
while CutMix cuts a patch from an image and pastes a new one from another image.
However, very few of existing studies focuses on graph-structured data.
In addition,
these methods
heavily depend on the information of soft labels
and cannot be directly applied in the unsupervised learning settings.

\begin{figure}[t]
\centering
\includegraphics[width=0.5\textwidth]{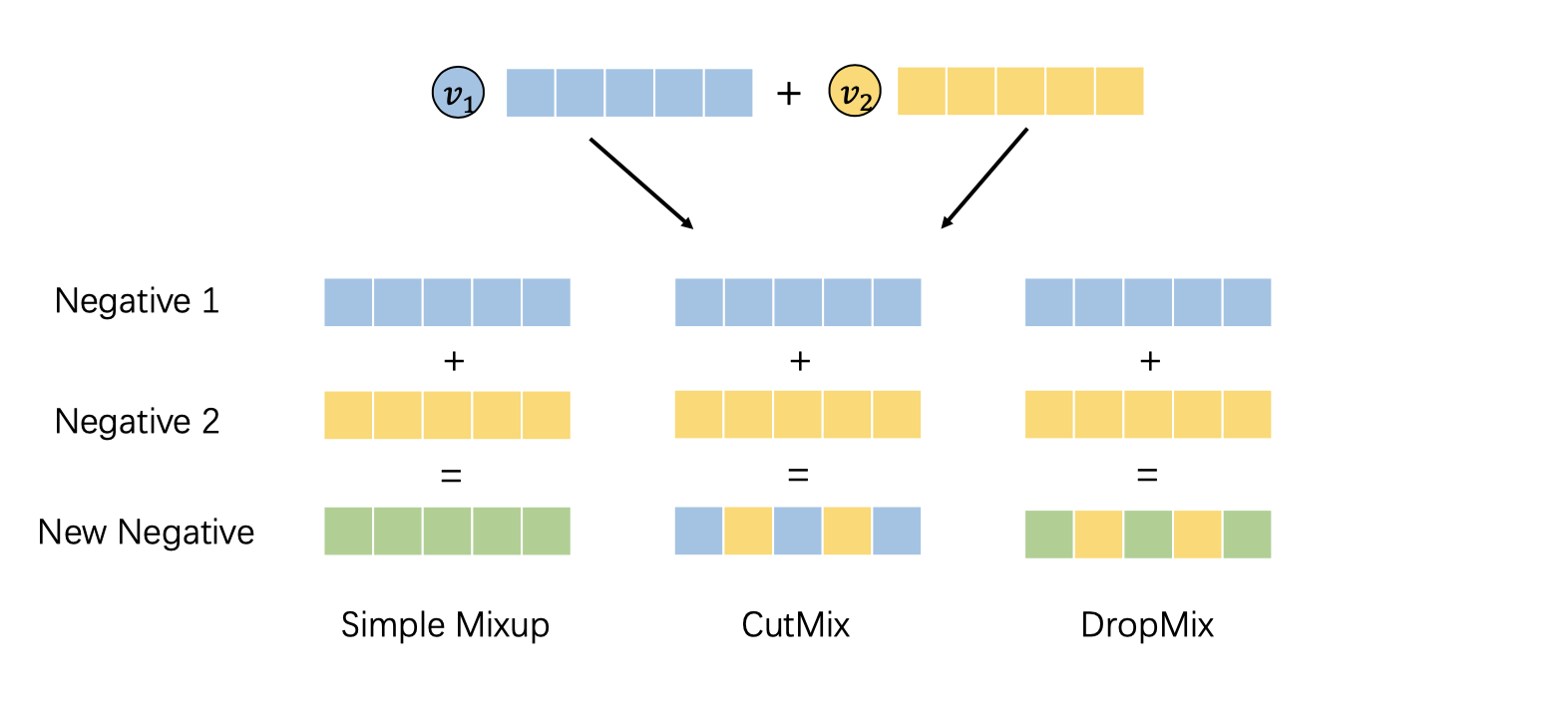}
\caption{
Three different Mix methods to synthesize harder negative samples.
Each square represents one dimension.
}
\label{fig1}
\end{figure}

To address these problems,
a recently proposed method ProGCL~\cite{xia2022progcl}
first selects hard negatives and then
uses Mixup~\cite{zhang2017mixup}
to generate harder ones in
graph contrastive learning (GCL),
which is essentially unsupervised learning.
Despite the success,
it has two major problems.
On the one hand,
it measures the hardness of negative samples based on only their local information in the graph,
while
the global information contained in the graph can further provide a supplementary view for 
hard negative selection.
On the other hand,
it performs Mixup on all the dimensions of hard negative representations.
However,
since the selected negative is already a hard one,
this could inadvertently lead to the information loss of the original sample
and adversely affect the quality of the newly synthesized negative.

In this paper, to tackle these issues, 
we propose a novel Mixup-based method {\model}
for synthesizing harder negative samples in GCL.
Our
goal is to
increase the hardness of negative samples 
without the supervision of soft labels
in the unsupervised learning setting,
while decreasing the information loss of the
original hard negative samples caused by Mixup. 
Specifically, 
we first measure the 
hardness
of negative samples by 
computing their similarities with positive samples from both local and global views of information. 
For the local view,
we consider the neighborhood of a node.
For the global view,
we employ
the diffusion matrix~\cite{klicpera2019diffusion} of the graph.
Then we rank the similarities and select hard negatives.
After that,
inspired by CutMix,
we mix these selected hard negatives 
only in partial dimensions to construct new harder negatives,
while keeping the remaining part unchanged.
This explains the origination of Drop in DropMix.
In this way,
we
can alleviate the information loss of the original hard negatives, even without the help of soft labels.
Figure~\ref{fig1} illustrates the difference between Mixup, CutMix, 
and our proposed method DropMix.
Finally, 
our main contributions in this paper are summarized as follows:
\begin{itemize}
    \item We present a novel method \model\ 
    for synthesizing harder negative samples 
    in GCL
    under the unsupervised learning setting.

    \item We unify the local and global views of information in the graph to
    measure the hardness of negative samples, which can be used to effectively filter easy negatives and false negatives.
    
    \item We propose to perform Mixup only on a proportion of dimensionalities of the hard negative samples, to decrease the information loss caused by Mixup.
    
     \item We conduct extensive experiments to show the superiority of our proposed method in generating harder negative samples in GCL.
     
\end{itemize}

\section{Related Work}

\subsection{Graph Contrastive Learning}
In recent years, inspired by contrastive learning in CV and NLP, more and more studies have begun to apply contrastive learning to graphs and have obtained inspiring results as well. 
As an unsupervised learning method, 
GCL has achieved comparable or even better results 
than many supervised learning methods,
without relying on the involvement of a large number of labels, which greatly improves the utilization efficiency of data in the real world.
For example,
a representative model DGI \cite{velickovic2019deep} first proposes to maximize the mutual information between node-level and graph-level representations to learn node representations. Based on DGI, MVGRL \cite{hassani2020contrastive} proposes to use node diffusion method to learn node representations by maximizing mutual information between node-level and graph-level representations of the original and diffusion graph.
Further,
GMI \cite{peng2020graph} develops an unsupervised learning model trained by maximizing the correlation between input graphs and high-level hidden representations. GraphCL \cite{you2020graph} explores the impact of different data augmentation methods on different domain datasets. BGRL \cite{thakoor2021bootstrapped} learns a node representation by encoding two augmented versions of a graph using two distinct graph encoders. Besides, MERIT \cite{jin2021multi} learns node representations by enhancing Siamese self-distillation with multi-scale contrastive learning. In addition, 
Grace \cite{zhu2020deep} and GCA \cite{Zhu_2021} are jointly consider corruption at both topology and node attribute levels to provide diverse contexts for nodes in different views.

\begin{figure*}[t]
\centering
\includegraphics[width=0.9\textwidth]{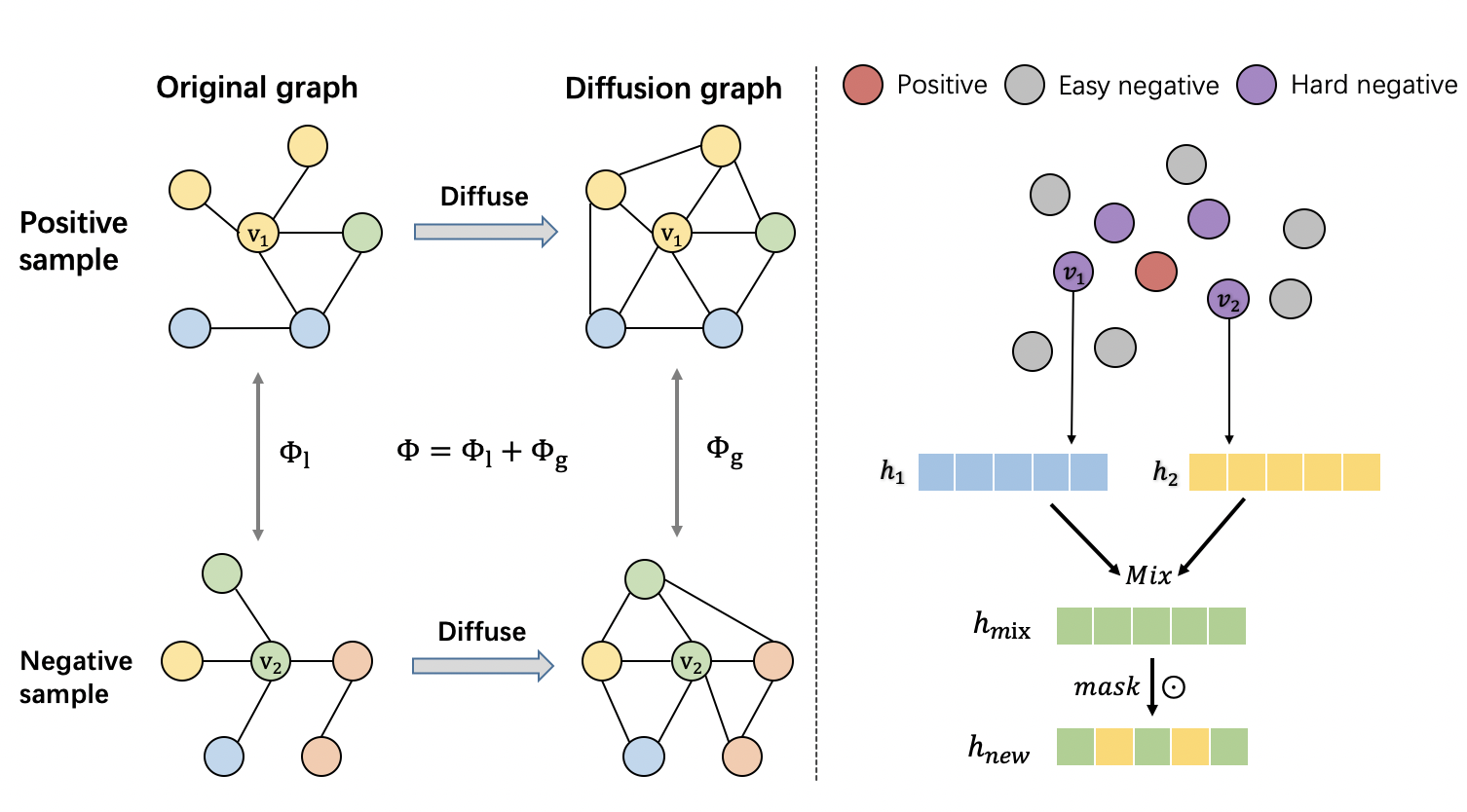}
\caption{(left) First step, we measure the hardness of negative samples in the original graph with local information and the diffusion graph with global information. The top and bottom graphs represent node $v_{1}$, $v_{2}$ and a subgraph structure around them respectively. (right) Second step, we construct new harder negative samples by mixing two hard negative samples in some dimensions. Specifically, $Mix$ and $\odot $ denote eq. \eqref{align 9} and mask respectively, which represent we mix $h_{1}$ and $h_{2}$ to generate green dimensions by mixing blue and yellow while retaining some yellow dimensions.}
\label{fig2}
\end{figure*}

\subsection{Negative Sampling}
Since 
the goal of contrastive learning is to pull the positive samples close to the anchor 
and push away the negative samples, how to construct positive and negative samples 
is particularly important. 
Some existing works 
use data augmentation methods~\cite{zhao2021data,you2020graph} to generate better and more representative positive samples, while the studies on generating negative samples are insufficient in comparison. 
Most recent studies on negative sampling 
focus on how to select better negative samples from negative sample sets. 
For example, 
HCL~\cite{robinson2020contrastive} 
first uses sampling distribution to generate negative samples
and then presents a sampling strategy on sampling in the absence of real dissimilar information.
What's more, KGPolicy~\cite{wang2020reinforced} considers both the similarities between the negative and the anchor (the positive samples).
After that, the two similarities are combined as the measure 
for selecting hard negative samples. 
For graph-structured data,
some works such as ~\cite{zhao2021graph} propose to cluster nodes 
in the 
training process, 
and generate pseudo-labels for nodes
based on the clustering results 
to decide which negative samples to select. 
Further, 
CuCo \cite{chu2021cuco} introduces Curriculum Learning and designs a scoring function for the negative samples, ranking them from easy to difficult and learning them sequentially.

\subsection{Mixup}
Mixup \cite{zhang2017mixup} is an effective method for image data augmentation.  
It interpolates between two images in proportion 
to generate 
a new sample and also
proportionally mixes their labels 
to get the corresponding soft label. Some recent studies have applied Mixup to graph data augmentation. 
For example, 
a recent work \cite{wang2021mixup} applies Mixup to both node classification and graph classification tasks with GNNs. GraphMix \cite{verma2021graphmix} proposes a data augmentation method for training fully connected networks jointly with GNNs through parameter sharing and interpolation-based regularization. Additionally, Mixup has also been used for the synthesis of hard negative samples recently. MoCHi \cite{kalantidis2020hard} first proposes to introduce Mixup into the construction of negative samples by mixing hard negative samples or positive samples with hard negative samples to synthesize harder negative samples artificially. And on graphs, ProGCL \cite{xia2022progcl} and M-Mix \cite{zhang2022m} similarly uses Mixup for the synthesis of hard negative samples.

\subsection{CutMix}
CutMix~\cite{yun2019cutmix} is another data augmentation method that combines both Dropout and Mixup. 
It first cuts a patch from an image and then pastes a new one from another image
to obtain new ones.
After that,
it mixes the ground truth labels proportionally according to the area size of the patch. 
In this way, 
CutMix can alleviate 
the problem that 
the samples generated by Mixup tend to be unnatural. 
However, this method is mainly developed for images but not for graph-structured data.

Despite the success, through experiments, 
we observe that the performances of hard negatives generated by Mixup and CutMix are not as good as we expected in the unsupervised learning settings.
Different from them, 
\model\ can 
generate 
better and harder negative samples without labels.

\section{Method}

\subsection{Problem Definition}
Let $G=(\mathcal{V}, \mathcal{E})$ denote an undirected graph, where $\mathcal{V}=\left \{v_{1},v_{2}...v_{N} \right \} $ is a set of $N$ nodes and $\mathcal{E}\in  {\mathcal{V}\times \mathcal{V}}$ denotes the adjacency relationships between nodes in $\mathcal{V}$. Additionally, node feature $ \mathcal{X}\in\mathbb{R}^{N\times d}$ is a d-dimensional matrix, and each node $v_{i}$ is associated with a feature vector $x_{i}$. The graph $G$ can be described as an adjacency matrix $A\in  \mathbb{R}^{N\times N}$ according to $\mathcal{E}$, so that we can compute the symmetrically normalized adjacency matrix $\hat{A}$ by $\hat{A}=\tilde{D}^{-\frac{1}{2}}\tilde{A}\tilde{D}^{-\frac{1}{2}}$, where $\tilde{A}$ is the adjacency matrix with added self-loops, and $\tilde{D}$ is the diagonal degree matrix of $\tilde{A}$.

\subsection{Graph Encoder}
\label{sec:gnn}
In this section, we use Graph Neutral Networks (GNNs) as our encoder to learn the nodes' representations through the message-passing scheme. For any node $v_{i}$ in the $l$-th layer, 
we can obtain its embedding by aggregating the information 
in the $(l-1)$-th layer from itself and all its neighbors:
\begin{equation}
    h_{u}^{(l)}=\texttt{AGGREGATE}\left ( h_{u}^{(l-1)}: u\in  \mathcal{N}(i) \right ) ,
\label{align 1}
\end{equation}%
\begin{equation}
h_{i}^{(l)}=\texttt{COMBINE}\left (h_{i}^{(l-1)}, h_{u}^{(l)} \right ) ,
\label{align 2}
\end{equation}%
where $h_{i}^{(l)}$ is the representation vector of node $v_{i}$ in the $l$-th layer with $h_{i}^{(0)}=x_{i}$. 
Further,
$\texttt{AGGREGATE}(\cdot)$ and $\texttt{COMBINE}(\cdot)$ are aggregation function and combination function of the GNN layer, respectively, and $\mathcal{N}(i)$ denotes the neighbor set of node $v_{i}$.

\subsection{Graph Contrastive Learning}
To achieve the goal of graph contrastive learning, i.e., minimizing the distance between the anchor and the positive samples,
while maximizing that between the anchor
and the negative samples,
we adopt the InfoNCE loss function \cite{oord2018representation}. 
Specifically, given a node $v_{i}$ and its positive sample $v_{j}$ which can be obtained by data augmentation, 
we can derive their embeddings $h_{i}$ and $h_{j}$ by GNN encoders.
And then
we take all the other nodes in the graph
as negative samples. 
Formally,
the training objective for the positive pair $(h_{i},h_{j})$ is defined as:
\begin{equation}
    L_{i,j}=-\texttt{log}\frac{\texttt{exp}(\texttt{sim}(h_{i},h_{j}/\tau ))}{\sum_{k=1}^{N}\texttt{exp}(\texttt{sim}(h_{i},h_{k}/\tau ))},
\label{align 3}
\end{equation}%
where $\tau$ denotes the temperature parameter and $\texttt{sim}(\cdot)$ represents the similarity function between $h_{i}$ and $h_{j}$.

\subsection{Hard Negative Selection}
In this section, we select hard negative samples from the negative sample set, which consists of nodes other than the positive.
And details are given in 
the left part of 
figure \ref{fig2}.
First, we consider that it may not enough to measure the degree of hardness of negative samples only from the local view of the original graph. Therefore, inspired by graph diffusion~\cite{klicpera2019diffusion}, we transform the adjacency matrix into a diffusion matrix, which can represent the global connections between two nodes.
After that, 
we measure the similarities between negative samples and 
positive samples from both the local and global views. 
Finally, we combine the two similarities to derive the measurement for the degree of hardness w.r.t. a given negative sample.

Similar as in~\cite{klicpera2019diffusion}, 
we define generalized graph diffusion that is computed by using fast approximation and sparsification methods in eq.~\eqref{align 4}.
Note that 
$T\in \mathbb{R}^{N\times N}$ is the generalized transition matrix and $\theta_{k}$ is the weighting coefficients, requiring $\sum_{k=0}^{\infty }\theta _{k}=1$, $\theta _{k}\in [0,1]$ and $\lambda _{i}\in [0,1]$, 
where $\lambda _{i}$ are eigenvalues of $T$.
\begin{equation}
    S = \sum_{k=0}^{\infty }\theta _{k}T^{k}.
\label{align 4}
\end{equation}%
In our work, we use Personalized PageRank (PPR) \cite{page1999pagerank}, which is one of the instantiations of generalized graph diffusion. 
Specifically, 
we set
$T=\hat{A}$ and $\theta _{k}=\alpha (1-\alpha )^{k}$, where $\alpha$ denotes teleport probability in a random walk. The closed-form solution to PPR diffusion is formulated as:
\begin{equation}
    S=\alpha (I_{n}-(1-\alpha )\hat{A})^{-1}.
\label{equation 5}
\end{equation}%
In this way, we have constructed a new  matrix $S$ that reflects the global relations between nodes in the graph.
Then 
we can measure the hardness of negative samples by 
computing their similarities
to the positive samples in the views of both the adjacency matrix and $S$. 
For simplicity,
we choose
cosine similarity as the measure function. 
Specifically, 
given a node $v$ and its positive sample $v_p$, 
for any negative sample $v_n$,
we use $\Phi_{l}$ and $\Phi_{g}$ to denote the hardness between $v_p$ and $v_n$ 
in the local and global views, respectively:
\begin{equation}
    \Phi_{l}=\frac{h_{p}^{T} h_{n}}{\left \| h_{p} \right \| \cdot \left \| h_{n} \right \|},
    \Phi_{g}=\frac{\tilde{h}_{p}^{T} \tilde{h}_{n} }{\left \| \tilde{h}_{p} \right \| \cdot \left \| \tilde{h}_{n} \right \|},
\label{align 6}
\end{equation}%
where $h_{p},{h_{n}},\tilde{h}_{p},\tilde{h}_{n}$ are the embeddings derived by the GNN encoder in section \ref{sec:gnn}. 
Here,
$h_{p},{h_{n}}$
denote the embeddings of positive and negative samples in the original graph, 
while
$\tilde{h}_{p},\tilde{h}_{n}$ represent that in the 
diffusion graph. 
After that, 
we combine the two measures 
to get the final degree of hardness $\Phi=\Phi_{l}+\Phi_{g}$ and then select hard negatives. 
Note that
the small hardness value generally corresponds to easy negative
 samples that are useless, 
 while the large one 
 indicates false negative samples that may hurt the model performance.
 Therefore, 
 we rank negative samples based on $\Phi$,
 remove samples at the two ends, and 
 select the remaining samples as hard negatives. 
 Specifically, we set two hyper-parameters $\alpha$ and $\beta$ to control the lower and upper limit of the hardness of hard negatives respectively.

\subsection{Hard Negative Mixing}
After selecting hard negative samples from the negative sample set,
we next 
mix the selected hard negatives 
to generate harder ones. 
The right part of figure \ref{fig2} shows a toy example on \model.
Let $h_{1}, h_{2}$ denote the representations of two hard negative samples, respectively. 
We first define the mixing operation for $h_{1}$ and $h_{2}$ as:
\begin{equation}
   h_{mix}=\lambda h_{1}+(1-\lambda )h_{2},
\label{align 9}
\end{equation}%
where $\lambda\in \left [ 0,1 \right ]$ is the mixing coefficient which decides the mixing ratio. 
Then, we mask a proportion of dimensions with a binary mask vector $M$ indicating which dimensions should be dropped out. In our work, we specify the proportion of one-valued entries in $M$ as a hyper-parameter to control the number of dimensions selected properly. Finally, we can obtain a new negative sample by combining eq. \eqref{align 9} and the mask operation:
\begin{equation}
  h_{new}=M\odot h_{1}+(1-M)\odot h_{mix}.
\label{align 10}
\end{equation}%
Based on eq.~\eqref{align 10}, 
we can 
generate new harder negative samples to help training with a novel mix method.

\section{Discussion}

\label{sec:discussion}
In this section,
we summarize the relationships between \model\ and two other data augmentation methods for hard negative generation: Mixup and CutMix.
First,
as shown in previous works \cite{zhang2017mixup,yun2019cutmix}
both Mixup and CutMix perform well in the supervised learning settings,
which require samples with labels. 
However,
in the unsupervised learning settings,
their performances could be degraded due to the lack of soft labels.
Further, 
Mixup performs interpolation over all the embedding dimensions of samples 
and could inadvertently lead to the information loss of the original hard negative sample.
For CutMix, 
although it can reduce the information loss by keeping some parts in the original negative unchanged, 
it discards other parts of the sample and loses the corresponding information completely.
Different from Mixup and CutMix,
\model\ 
mixes the representations 
of samples only in partial dimensions
and keeps others unchanged,
which decreases the information loss in the original negative sample.
This
further weakens the need on soft labels and extends its applicability in the unsupervised learning settings.
Technically,
\model\ subsumes both Mixup and CutMix.

\begin{table}[t]

    \caption{Statistics of datasets}
    \label{table 1}
    \centering
    \tabcolsep=0.2cm
    \renewcommand\arraystretch{1.2}
    \begin{tabular}{ccccc}   
        \toprule
         Datasets  & Nodes & Edges & Attributes & Classes   \\
        \midrule
         Cora     & 2,708 & 5,429 & 1,433 & 7     \\
         Citeseer  & 3,327 & 4,732 & 3,703 & 6    \\
         Pubmed     & 19,717 & 44,338 & 500 & 3        \\
         Wiki-CS     & 11,701 & 216,123 & 300 & 10      \\
         Amazon-Photo   & 7,650 & 119,081 & 745 & 8      \\         Coauthor-CS    & 18,333 & 81,894 & 6,805 & 15  \\
        \bottomrule
    \end{tabular}
\end{table}

\begin{table*}
    \caption{ The classification accuracy (\%) over the methods on 6 datasets. ``OOM" denotes out of memory on a 32GB GPU. 
    The highest performances are highlighted in bold.
    }
    \label{table 2}
    \centering
    \tabcolsep=0.3cm
    \renewcommand\arraystretch{1.4}
    \scalebox{0.95}{
    \begin{tabular}{ccccccc} 
        \toprule
        Methods  & Cora & Citeseer & Pubmed & Wiki-CS  &  Amazon-Photo & Coauthor-CS   \\
        \midrule
         GCN     & 81.80 $\pm$ 0.50 & 70.80 $\pm$ 0.50 & 79.30 $\pm$ 0.70 & 77.19 $\pm$ 0.12 &  92.42 $\pm$ 0.22 & 93.03 $\pm$ 0.31     \\
         GAT    & 83.00 $\pm$ 0.70 & 72.50 $\pm$ 0.70 & 79.00 $\pm$ 0.30 & 77.65 $\pm$ 0.11  &  92.56 $\pm$ 0.35 & 92.31 $\pm$ 0.24     \\  
        \midrule
         GAE     & 71.55 $\pm$ 0.33 & 65.87 $\pm$ 0.42 & 72.15 $\pm$ 0.50 & 70.15 $\pm$ 0.01 &  91.62 $\pm$ 0.13 & 90.01 $\pm$ 0.17     \\
         VGAE    & 73.27 $\pm$ 0.47 & 66.92 $\pm$ 0.50 & 74.13 $\pm$ 0.62 & 75.35 $\pm$ 0.14  &  92.20 $\pm$ 0.11 & 92.11 $\pm$ 0.09     \\
         DGI     & 83.80 $\pm$ 0.50 & 72.00 $\pm$ 0.60 & 77.90 $\pm$ 0.30 & 75.35 $\pm$ 0.14  &  91.61 $\pm$ 0.22 & 92.15 $\pm$ 0.63       \\
         GMI     & 83.00 $\pm$ 0.30 & 72.40 $\pm$ 0.10 & 79.90 $\pm$ 0.20 & 74.85 $\pm$ 0.08  &  90.68 $\pm$ 0.17 & OOM       \\
         MVGRL   & 86.80 $\pm$ 0.50 & 73.30 $\pm$ 0.50 & 80.10 $\pm$ 0.70 & 77.43 $\pm$ 0.17  &  92.08 $\pm$ 0.01 & 92.18 $\pm$ 0.05       \\
         BGRL    & 84.68 $\pm$ 0.23 & 73.88 $\pm$ 0.15 & 80.73 $\pm$ 0.17 & 78.41 $\pm$ 0.09  &  92.95 $\pm$ 0.07 & 92.72 $\pm$ 0.03       \\
         MERIT   & 83.10 $\pm$ 0.60 & 74.00 $\pm$ 0.70 & 80.10 $\pm$ 0.40 & 78.35 $\pm$ 0.05  &  92.53 $\pm$ 0.15 & 92.51 $\pm$ 0.14       \\
         ProGCL  & 83.50 $\pm$ 0.22 & 74.19 $\pm$ 0.13 & 80.77 $\pm$ 0.15 & 78.45 $\pm$ 0.04  &  93.64 $\pm$ 0.13 & 93.67 $\pm$ 0.12     \\
        \midrule
        MVGRL+Mixup    & 86.92 $\pm$ 0.44 & 74.08 $\pm$ 0.28 & 80.62 $\pm$ 0.52 & 78.23 $\pm$ 0.46  &  93.55 $\pm$ 0.63 & 94.66 $\pm$ 0.21        \\
        MVGRL+CutMix   & 86.87 $\pm$ 0.66 & 74.24 $\pm$ 0.52 & 80.68 $\pm$ 0.40 & 78.42 $\pm$ 0.64  &  93.47 $\pm$ 0.39 & 95.21 $\pm$ 0.50       \\
        MVGRL+\model & \pmb{87.17 $\pm$ 0.31} & \pmb{74.74 $\pm$ 0.23} & \pmb{81.29 $\pm$ 0.26} & \pmb{78.82 $\pm$ 0.16}  &  \pmb{94.46 $\pm$ 0.33} & \pmb{96.66 $\pm$ 0.52}       \\
        
        \bottomrule
    \end{tabular}
    }
\end{table*}

\section{Experiments}
In this section, we evaluate DropMix's effectiveness. We compare DropMix with 10 other methods by their accuracies on node classification tasks. 
We also analyze the performance of each component of \model\ by ablation study and hyper-parameter sensitivity analysis. Further, 
we verify that \model\ can be used in different GCL models and it is very suitable for graph-structured data.

\subsection{Datasets}
We use six datasets which are widely used for node classification tasks to verify the performance of DropMix including Cora, Citeseer, Pubmed, Amazon-Photo, Wiki-CS and Coauthor-CS. The first three are citation networks, where each node represents a document and each edge is a citation link. Amazon Photo is a segment of the Amazon co-purchase graph, where nodes represent goods and edges indicate that two goods are frequently bought together. Wiki-CS is a Wikipedia-based dataset, which consists of nodes corresponding to Computer Science articles, with edges based on hyperlinks and 10 classes representing different branches of the field. Coauthor CS is a co-authorship graph based on the Microsoft Academic Graph from the KDD Cup 2016 challenge. Here, nodes are authors, which are connected by an edge if they co-authored a paper, node features represent paper keywords for each author’s papers, and class labels indicate most active fields of study for each author. The specific statistics are summarized in Table \ref{table 1}.

\subsection{Experimental Setting}
We implement DropMix using PyTorch 1.10.0. We adopt the task of node classification to evaluate the performance of node representation learning. 
Further, We adopt MVGRL \cite{hassani2020contrastive} as the base model and use test accuracy as the evaluation indicator to demonstrate the effectiveness of DropMix. We share the GNN encoder in the two views of local and global. 
For DropMix and all its variants, we set the learning rate to 0.001 on Cora, Citeseer and Pubmed, 0.0007 on the other, and the penalty weight on the $l_{2}$-norm regularizer to 0.002 on Citeseer. We use early stopping with the patience of 40 on Amazon-photo, Coauthor-CS and Wiki, and 20 on the other datasets, i.e., we stop training if the validation accuracy does not decrease for 40 or 20 consecutive epochs. 
We fine-tune the model hyper-parameters by grid search.
Specially, we set $\gamma$ as the rate of representation dimensions \{0.1, 0.2, 0.3, 0.4, 0.5, 0.6\}, and the mixup rate $\lambda$ \{0.1, 0.2, 0.3, 0.4\}. 
What's more, to the lower limit proportion of the hardness of hard negatives, we fine-tune it from 5\% to 50\% with a 5\% increment each time. 
And to the upper limit proportion, we fine-tune it from \{80\%, 85\%, 90\%, 95\%\}.
For a fair comparison, we run all the experiments on a server with 32G memory and a single NVIDIA 2080Ti GPU. 
Additionally, for all the baseline methods, we use the original results released by their papers, and for some methods whose other experimental results on some datasets can not be found, we use the records which are reported in \cite{xia2022progcl} and \cite{jin2021multi}. However, for the latest work ProGCL, we use the default parameters reported in the original paper and the original codes to run the experiment on Cora, Citeseer and Pubmed. For each method, we run experiments 10 times and report the average results.

\begin{figure*}[t]
\centering
\includegraphics[width=0.95\textwidth]{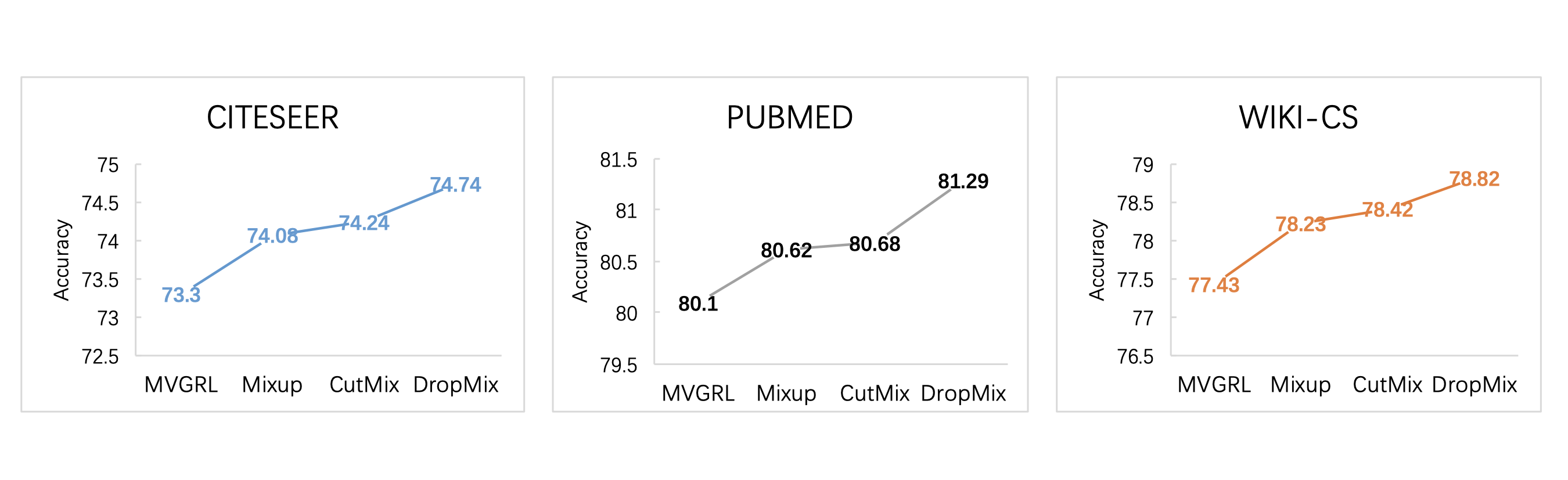}
\caption{The results of different mix methods on three datasets.}
\label{fig3}
\end{figure*}

\subsection{Comparison with the State-of-the-Arts}
To demonstrate the effectiveness of DropMix, we first evaluate it with 10 state-of-the-art methods which can be grouped into two categories on six datasets. Specifically, we choose multiple baselines which are unsupervised learning methods including Graph AutoEncoders(GAE, VGAE) \cite{kipf2016variational}, DGI \cite{velickovic2019deep}, GMI \cite{peng2020graph}, MVGRL \cite{hassani2020contrastive}, BGRL \cite{thakoor2021bootstrapped}, MERIT \cite{jin2021multi} and ProGCL \cite{xia2022progcl} which is the most advanced method introducing Mixup into the synthesis of negative samples as introduced in the related work. 
What's more, we also compare \model\ with supervised learning methods including GCN \cite{kipf2016semi}, 
and Graph Attention Network (GAT). GCN is a model that extends the convolution operation on graphs.
And Graph Attention Network further integrates the attention mechanism in the convolutional layer. 
In addition, we use only Mixup or CutMix, respectively, in the same unsupervised learning setting to generate new negative samples as the methods we compared.

Table \ref{table 2} summarizes the performance results and the best results are in bold. 
We can observe that the test accuracy of DropMix outperforms all the other methods on six datasets. 
This indicates that we can improve the performance of a GCL model significantly only by supplying high-quality negative samples.
Compared with ProGCL, which also uses Mixup to construct new negative samples, DropMix selects better hard negative samples by considering local information and global information simultaneously.
Furthermore, it can be seen from the table that, although Mixup and CutMix can both bring some degree of improvement to the base model, the performance of DropMix is better.
This is because it only mixes parts of dimensions to generate new negative samples so that less original information of the hard negative samples can be lost.

\begin{table}
    \centering
    \tabcolsep=0.12cm
    \renewcommand\arraystretch{1.3}
    \caption{The results of different methods based GCA
    , the best performances are highlighted in bold.
    }
    \label{table 3}
    \begin{tabular}{cccc}   
        \toprule
         Datasets  & Cora  &  Citeseer  & Amazon-Photo\\
        \midrule
         GCA  &  80.90 $\pm$ 0.41  & 72.14 $\pm$ 0.06   &  92.55 $\pm$ 0.03  \\
         ProGCL  &  83.50 $\pm$ 0.22  & 74.19 $\pm$ 0.13  &  93.64 $\pm$ 0.13   \\
         GCA+DropMix & \pmb{83.82 $\pm$ 0.25} & \pmb{74.95 $\pm$ 0.20}  &  \pmb{94.13 $\pm$ 0.11} \\
        \bottomrule
    \end{tabular}
\end{table}

\begin{table}
    \caption{The ablation study of multi-view measure on three datasets
    , the best performances are highlighted in bold.
    }
    \label{table 4}
    \centering
    \tabcolsep=0.18cm
    \renewcommand\arraystretch{1.3}
    \begin{tabular}{cccc}   
        \toprule
        Datasets  & Wiki-CS & Amazon-Photo & Coauthor-CS   \\
        \midrule
        \model-ol    & 78.53 $\pm$ 0.36 & 94.19 $\pm$ 0.55 & 95.62 $\pm$ 0.28    \\
        \model-og   & 78.71 $\pm$ 0.43 & 94.08 $\pm$ 0.30 & 96.32 $\pm$ 0.39  \\
        \model   & \pmb{78.82 $\pm$ 0.16} & \pmb{94.46 $\pm$ 0.33} & \pmb{96.66 $\pm$ 0.52}   \\
    
        \bottomrule
    \end{tabular}
\end{table}

\subsection{Results on different GCL Models}

We evaluate the performance of \model\ not only on the method of MVGRL, but on GCA \cite{Zhu_2021} to get a fairer comparison with ProGCL which is based on GCA. Results are shown in Table 
\ref{table 3}. The experimental results of GCA are taken from \cite{xia2022progcl} and \cite{liu2021rethinking}. It can be seen from the table that the performance is better than the base model GCA by adding DropMix on it. Besides this can demonstrate that DropMix can be used in different GCL methods to improve their performances by reducing information loss when constructing new hard negatives.

\subsection{Ablation study}
\label{sec: ablation study}
In this section, we conduct an ablation study on DropMix to further understand the characteristics of its main components. 
We consider different variants of the two steps to study the effect of multi-views measure and partial dimensions mixing respectively.

\subsubsection{Effect of multi-views measure}
In our method, how to select hard negatives is important. We consider that measuring the hardness of negative samples from both the local view and global view may lead to better performance. 
Therefore, we do the same operation as \model\ for three datasets, but only use local or global information respectively to measure.
The results are shown in Table \ref{table 4}, where ``\model-ol'' means DropMix with only local view and ``\model-og'' means DropMix with only global view. 
We can observe that DropMix achieves better performance than both DropMix-ol and DropMix-og, and DropMix-og beats DropMix-ol in most cases. This 
shows that compared with only measuring the hardness by local information, global information may show better performance, because of the rich information it incorporates from the global view.
However, it's obvious that the best way to measure is considering the two views at the same time.

\subsubsection{Effect of partial dimensions mixing}
We replace our method with the ordinary Mixup and CutMix on node representations to study the impact of partial dimensions mixing.
Mixup denotes we mix all the dimensions at the same time, and CutMix denotes we change some dimensions of a negative sample by the corresponding dimensions of another one.
We evaluate the performance of the three methods and report the performance in Figure \ref{fig3}. 
We can observe that Mixup and CutMix can bring some degree of improvement. 
Nevertheless, DropMix clearly outperforms them additionally on three datasets.
This indicates that our method which mixes node representations on partial dimensions achieves success in GCL, and further shows the importance of reducing hard information loss while generating new hard negatives.

\begin{table}
    \caption{The results of different methods based MVGRL
    , and the highest performances are marked in bold. 
    }
    \label{table 5}
    \centering
    \tabcolsep=0.1cm
    \renewcommand\arraystretch{1.22}
    \begin{tabular}{cccc}   
        \toprule
         Datasets  &  Citeseer  & Amazon-Photo\\
        \midrule
         MVGRL  &  73.30 $\pm$ 0.50 &  92.08 $\pm$ 0.01  \\
         \midrule
         Input Feature + Mixup  &  73.60 $\pm$ 0.78  &  92.82 $\pm$ 0.45  \\
         Input Feature + CutMix  &  73.89 $\pm$ 0.66  &  92.56 $\pm$ 0.30  \\
         Input Feature + DropMix  &  73.82 $\pm$ 0.42  &  92.76 $\pm$ 0.50  \\
         \midrule
         Node Representation + Mixup  &  74.08 $\pm$ 0.19  &  93.55 $\pm$ 0.41  \\
         Node Representation + CutMix  &  74.24 $\pm$ 0.37  &  93.47 $\pm$ 0.42  \\
         Node Representation + DropMix  &  \pmb{74.74 $\pm$ 0.23}  &  \pmb{94.46 $\pm$ 0.33}  \\
        \bottomrule
    \end{tabular}
    
\end{table}

\begin{figure}[t]
\centering
\includegraphics[width=0.5\textwidth]{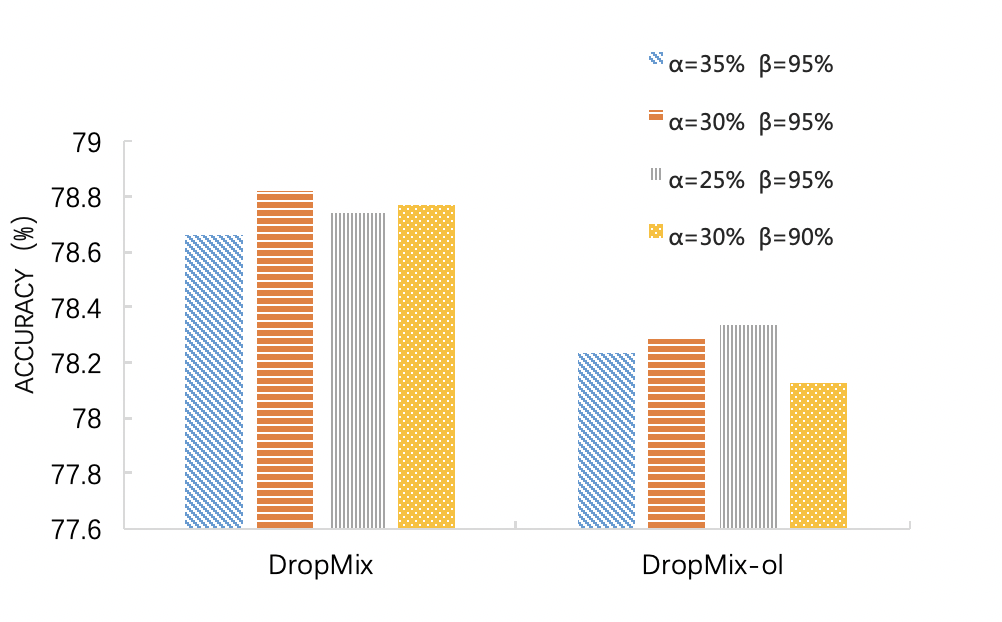}
\caption{Parameter analysis of the hardness of hard negative samples on Wiki-CS.}
\label{fig4}
\end{figure}

\subsection{Effect of DropMix on Input Feature and Node Representation}
Both Mixup and CutMix are data augmentation methods that are usually used in the input layer. To analyze the performance of DropMix on graph-structured data to synthesize negative samples, we show the results of DropMix and others implemented in the input layer and node representation generated by GNN, respectively. The results are shown in Table \ref{table 5}.
It can be seen that the performance of mixing on node representations is better than on the input feature, this may benefit from the graph structure information used in GNN. Further, we observe that in the input layer, the performance gap of DropMix against Mixup and CutMix is marginal, and even worse because of the sparse characteristic of graph data. However, when we use our method on node representations, DropMix presents its superiority on graph-structured data.

\subsection{Hyper-parameters Sensitivity Analysis}
We end this section with a sensitivity analysis of the hyper-parameters. In particular, we study four key hyper-parameters in two steps: The lower and upper hardness limit $\alpha$ and $\beta$ of hard negatives we select in the first step, the rate $\gamma$ of dimensions to mix, and the rate $\lambda $ of mixup. In our experiments, we vary one parameter each time with others fixed.

\begin{figure}[t]
\centering
\includegraphics[width=0.5\textwidth]{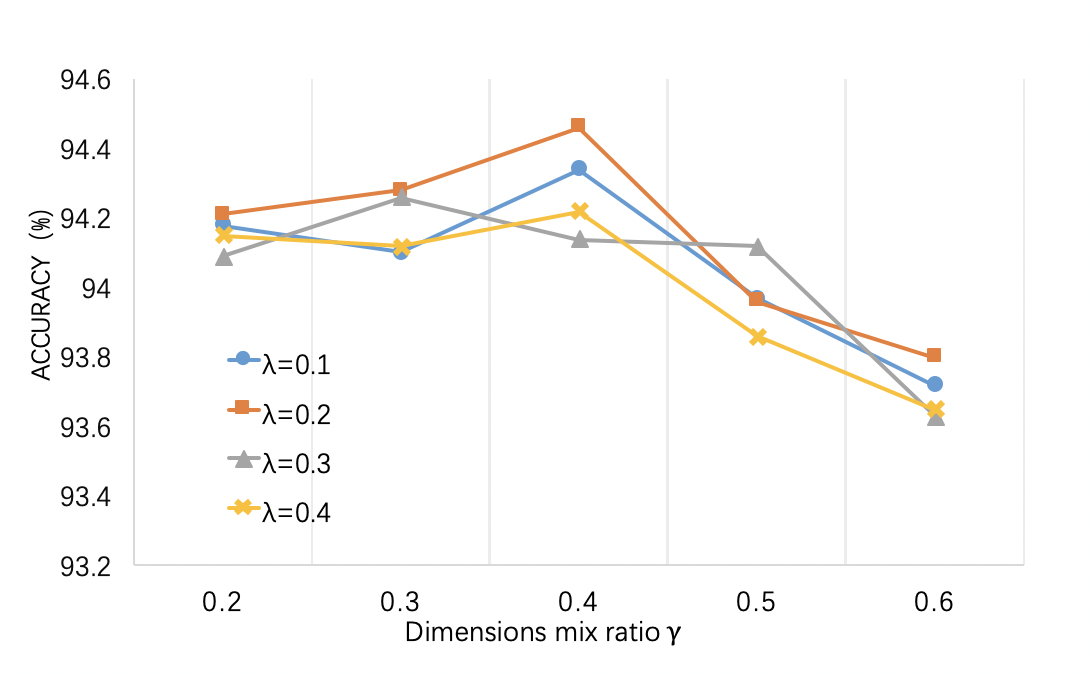}
\caption{Parameter analysis of mix on Amazon-photo.}
\label{fig5}
\end{figure}

\subsubsection{Number of hard negative samples}	

In section \ref{sec: ablation study}, we have studied the effect of multi-views measurement. 
However, 
after measuring the hardness of negative samples, 
how to choose the most proper hardness range is also important.
We use hyper-parameters $\alpha$ and $\beta$ to control the lower and upper limit of the range, respectively. 
For example, $\alpha$=35\% and $\beta$=95\% 
denote that we choose hard negative samples whose hardness values are ranked within 35\%-95\% among all the negative samples.
And then, we use the most proper $\alpha$ and $\beta$ in both DropMix and DropMix-ol to achieve the best performance respectively, where ``DropMix-ol'' means only using local information. 
Figure \ref{fig4} shows the results on Wiki-CS. 
We can observe that too high or too low hardness both hurt the performance so it's essential to discard easy negatives that are useless and the hardest ones which may be false negatives.
Additionally, 
it can be seen that 
through measuring from multi-views, we can not only select more precise hard negatives but get better results through fewer negatives.

\subsubsection{Hyper-parameters of mix}

We show the results of the two hyper-parameters on Amazon-Photo in Figure \ref{fig5}, where the scores on the horizontal axis denote the number of dimensions we randomly selected as a proportion of all dimensions. 
From this figure, we can see that both hyper-parameters can impact the performance of the method, where the rate of dimensions to mixup is more important. 
It's obvious that if $\gamma$ is more than 0.5, the performance drops rapidly. Thus, it can prove that it's necessary to decrease the information loss of the original negatives. 
However, a too small $\gamma$ may lead to the mix being ineffective. This further shows that Dropmix which mixes on partial dimensions is an effective method for generating harder negative samples.

\section{Conclusion}
In this paper, we are devoted to learning better representations for GCL. We propose a novel method to synthesize harder negative samples.
Specifically, we present a two-step method to implement. 
We first select some hard negative samples by measuring their hardness from both local and global views in the graph simultaneously. 
Second, we mix hard negative samples only on partial dimensions to increase the hardness of negative samples and decrease the information loss caused by Mixup.
Finally, experimental results demonstrate that our method performs favorably on six datasets.


\section*{Acknowledgment}
This work is supported by Shanghai Pujiang Talent Program No. 21PJ1402900, Shanghai Science and Technology Committee General Program No. 22ZR1419900 and National Natural Science Foundation of China No. 62202172.

\bibliographystyle{IEEEtran}
\bibliography{ref}

\begin{thebibliography}{10}
\providecommand{\url}[1]{#1}
\csname url@samestyle\endcsname
\providecommand{\newblock}{\relax}
\providecommand{\bibinfo}[2]{#2}
\providecommand{\BIBentrySTDinterwordspacing}{\spaceskip=0pt\relax}
\providecommand{\BIBentryALTinterwordstretchfactor}{4}
\providecommand{\BIBentryALTinterwordspacing}{\spaceskip=\fontdimen2\font plus
\BIBentryALTinterwordstretchfactor\fontdimen3\font minus
  \fontdimen4\font\relax}
\providecommand{\BIBforeignlanguage}[2]{{%
\expandafter\ifx\csname l@#1\endcsname\relax
\typeout{** WARNING: IEEEtran.bst: No hyphenation pattern has been}%
\typeout{** loaded for the language `#1'. Using the pattern for}%
\typeout{** the default language instead.}%
\else
\language=\csname l@#1\endcsname
\fi
#2}}
\providecommand{\BIBdecl}{\relax}
\BIBdecl

\bibitem{neville2000iterative}
J.~Neville and D.~Jensen, ``Iterative classification in relational data,'' in
  \emph{Proc. AAAI-2000 workshop on learning statistical models from relational
  data}, 2000, pp. 13--20.

\bibitem{zhu2003semi}
X.~Zhu, Z.~Ghahramani, and J.~D. Lafferty, ``Semi-supervised learning using
  gaussian fields and harmonic functions,'' in \emph{Proceedings of the 20th
  International conference on Machine learning (ICML-03)}, 2003, pp. 912--919.

\bibitem{bachman2019learning}
P.~Bachman, R.~D. Hjelm, and W.~Buchwalter, ``Learning representations by
  maximizing mutual information across views,'' \emph{Advances in neural
  information processing systems}, vol.~32, 2019.

\bibitem{he2020momentum}
K.~He, H.~Fan, Y.~Wu, S.~Xie, and R.~Girshick, ``Momentum contrast for
  unsupervised visual representation learning,'' in \emph{Proceedings of the
  IEEE/CVF conference on computer vision and pattern recognition}, 2020, pp.
  9729--9738.

\bibitem{chen2020simple}
T.~Chen, S.~Kornblith, M.~Norouzi, and G.~Hinton, ``A simple framework for
  contrastive learning of visual representations,'' in \emph{International
  conference on machine learning}.\hskip 1em plus 0.5em minus 0.4em\relax PMLR,
  2020, pp. 1597--1607.

\bibitem{devlin2018bert}
J.~Devlin, M.-W. Chang, K.~Lee, and K.~Toutanova, ``Bert: Pre-training of deep
  bidirectional transformers for language understanding,'' \emph{arXiv preprint
  arXiv:1810.04805}, 2018.

\bibitem{yang2019xlnet}
Z.~Yang, Z.~Dai, Y.~Yang, J.~Carbonell, R.~R. Salakhutdinov, and Q.~V. Le,
  ``Xlnet: Generalized autoregressive pretraining for language understanding,''
  \emph{Advances in neural information processing systems}, vol.~32, 2019.

\bibitem{kipf2016semi}
T.~N. Kipf and M.~Welling, ``Semi-supervised classification with graph
  convolutional networks,'' \emph{arXiv preprint arXiv:1609.02907}, 2016.

\bibitem{chen2020improved}
X.~Chen, H.~Fan, R.~Girshick, and K.~He, ``Improved baselines with momentum
  contrastive learning,'' \emph{arXiv preprint arXiv:2003.04297}, 2020.

\bibitem{tian2020makes}
Y.~Tian, C.~Sun, B.~Poole, D.~Krishnan, C.~Schmid, and P.~Isola, ``What makes
  for good views for contrastive learning?'' \emph{Advances in Neural
  Information Processing Systems}, vol.~33, pp. 6827--6839, 2020.

\bibitem{zhao2021data}
T.~Zhao, Y.~Liu, L.~Neves, O.~Woodford, M.~Jiang, and N.~Shah, ``Data
  augmentation for graph neural networks,'' in \emph{Proceedings of the AAAI
  Conference on Artificial Intelligence}, vol.~35, no.~12, 2021, pp.
  11\,015--11\,023.

\bibitem{you2020graph}
Y.~You, T.~Chen, Y.~Sui, T.~Chen, Z.~Wang, and Y.~Shen, ``Graph contrastive
  learning with augmentations,'' \emph{Advances in Neural Information
  Processing Systems}, vol.~33, pp. 5812--5823, 2020.

\bibitem{cai2020all}
T.~T. Cai, J.~Frankle, D.~J. Schwab, and A.~S. Morcos, ``Are all negatives
  created equal in contrastive instance discrimination?'' \emph{arXiv preprint
  arXiv:2010.06682}, 2020.

\bibitem{robinson2020contrastive}
J.~Robinson, C.-Y. Chuang, S.~Sra, and S.~Jegelka, ``Contrastive learning with
  hard negative samples,'' \emph{arXiv preprint arXiv:2010.04592}, 2020.

\bibitem{wang2020reinforced}
X.~Wang, Y.~Xu, X.~He, Y.~Cao, M.~Wang, and T.-S. Chua, ``Reinforced negative
  sampling over knowledge graph for recommendation,'' in \emph{Proceedings of
  the web conference 2020}, 2020, pp. 99--109.

\bibitem{zhang2017mixup}
H.~Zhang, M.~Cisse, Y.~N. Dauphin, and D.~Lopez-Paz, ``mixup: Beyond empirical
  risk minimization (2017),'' \emph{arXiv preprint arXiv:1710.09412}, 2017.

\bibitem{yun2019cutmix}
S.~Yun, D.~Han, S.~J. Oh, S.~Chun, J.~Choe, and Y.~Yoo, ``Cutmix:
  Regularization strategy to train strong classifiers with localizable
  features,'' in \emph{Proceedings of the IEEE/CVF international conference on
  computer vision}, 2019, pp. 6023--6032.

\bibitem{xia2022progcl}
J.~Xia, L.~Wu, G.~Wang, J.~Chen, and S.~Z. Li, ``Progcl: Rethinking hard
  negative mining in graph contrastive learning,'' in \emph{International
  Conference on Machine Learning}.\hskip 1em plus 0.5em minus 0.4em\relax PMLR,
  2022, pp. 24\,332--24\,346.

\bibitem{klicpera2019diffusion}
J.~Klicpera, S.~Wei{\ss}enberger, and S.~G{\"u}nnemann, ``Diffusion improves
  graph learning,'' \emph{arXiv preprint arXiv:1911.05485}, 2019.

\bibitem{velickovic2019deep}
P.~Velickovic, W.~Fedus, W.~L. Hamilton, P.~Li{\`o}, Y.~Bengio, and R.~D.
  Hjelm, ``Deep graph infomax.'' \emph{ICLR (Poster)}, vol.~2, no.~3, p.~4,
  2019.

\bibitem{hassani2020contrastive}
K.~Hassani and A.~H. Khasahmadi, ``Contrastive multi-view representation
  learning on graphs,'' in \emph{International Conference on Machine
  Learning}.\hskip 1em plus 0.5em minus 0.4em\relax PMLR, 2020, pp. 4116--4126.

\bibitem{peng2020graph}
Z.~Peng, W.~Huang, M.~Luo, Q.~Zheng, Y.~Rong, T.~Xu, and J.~Huang, ``Graph
  representation learning via graphical mutual information maximization,'' in
  \emph{Proceedings of The Web Conference 2020}, 2020, pp. 259--270.

\bibitem{thakoor2021bootstrapped}
S.~Thakoor, C.~Tallec, M.~G. Azar, R.~Munos, P.~Veli{\v{c}}kovi{\'c}, and
  M.~Valko, ``Bootstrapped representation learning on graphs,'' in \emph{ICLR
  2021 Workshop on Geometrical and Topological Representation Learning}, 2021.

\bibitem{jin2021multi}
M.~Jin, Y.~Zheng, Y.-F. Li, C.~Gong, C.~Zhou, and S.~Pan, ``Multi-scale
  contrastive siamese networks for self-supervised graph representation
  learning,'' \emph{arXiv preprint arXiv:2105.05682}, 2021.

\bibitem{zhu2020deep}
Y.~Zhu, Y.~Xu, F.~Yu, Q.~Liu, S.~Wu, and L.~Wang, ``Deep graph contrastive
  representation learning,'' \emph{arXiv preprint arXiv:2006.04131}, 2020.

\bibitem{Zhu_2021}
------, ``Graph contrastive learning with adaptive augmentation,'' in
  \emph{Proceedings of the Web Conference 2021}, 2021, pp. 2069--2080.

\bibitem{zhao2021graph}
H.~Zhao, X.~Yang, Z.~Wang, E.~Yang, and C.~Deng, ``Graph debiased contrastive
  learning with joint representation clustering.'' in \emph{IJCAI}, 2021, pp.
  3434--3440.

\bibitem{chu2021cuco}
G.~Chu, X.~Wang, C.~Shi, and X.~Jiang, ``Cuco: Graph representation with
  curriculum contrastive learning.'' in \emph{IJCAI}, 2021, pp. 2300--2306.

\bibitem{wang2021mixup}
Y.~Wang, W.~Wang, Y.~Liang, Y.~Cai, and B.~Hooi, ``Mixup for node and graph
  classification,'' in \emph{Proceedings of the Web Conference 2021}, 2021, pp.
  3663--3674.

\bibitem{verma2021graphmix}
V.~Verma, M.~Qu, K.~Kawaguchi, A.~Lamb, Y.~Bengio, J.~Kannala, and J.~Tang,
  ``Graphmix: Improved training of gnns for semi-supervised learning,'' in
  \emph{Proceedings of the AAAI Conference on Artificial Intelligence},
  vol.~35, no.~11, 2021, pp. 10\,024--10\,032.

\bibitem{kalantidis2020hard}
Y.~Kalantidis, M.~B. Sariyildiz, N.~Pion, P.~Weinzaepfel, and D.~Larlus, ``Hard
  negative mixing for contrastive learning,'' \emph{Advances in Neural
  Information Processing Systems}, vol.~33, pp. 21\,798--21\,809, 2020.

\bibitem{zhang2022m}
S.~Zhang, M.~Liu, J.~Yan, H.~Zhang, L.~Huang, X.~Yang, and P.~Lu, ``M-mix:
  Generating hard negatives via multi-sample mixing for contrastive learning,''
  in \emph{Proceedings of the 28th ACM SIGKDD Conference on Knowledge Discovery
  and Data Mining}, 2022, pp. 2461--2470.

\bibitem{oord2018representation}
A.~v.~d. Oord, Y.~Li, and O.~Vinyals, ``Representation learning with
  contrastive predictive coding,'' \emph{arXiv preprint arXiv:1807.03748},
  2018.

\bibitem{page1999pagerank}
L.~Page, S.~Brin, R.~Motwani, and T.~Winograd, ``The pagerank citation ranking:
  Bringing order to the web.'' Stanford InfoLab, Tech. Rep., 1999.

\bibitem{kipf2016variational}
T.~N. Kipf and M.~Welling, ``Variational graph auto-encoders,'' \emph{arXiv
  preprint arXiv:1611.07308}, 2016.

\bibitem{liu2021rethinking}
Z.~Liu, H.~Feng, and C.~Wang, ``Rethinking temperature in graph contrastive
  learning,'' 2021.

\end{thebibliography}

\end{document}